\documentclass[pmlr,twocolumn,10pt]{jmlr} 

\usepackage{booktabs}
\usepackage{siunitx}

\usepackage[switch]{lineno}

\theorembodyfont{\upshape}
\theoremheaderfont{\scshape}
\theorempostheader{:}
\theoremsep{\newline}

\jmlrvolume{333}
\jmlryear{2026}
\jmlrsubmitted{LEAVE UNSET}
\jmlrpublished{LEAVE UNSET}
\jmlrworkshop{Conference on Health, Inference, and Learning (CHIL) 2026} 

\title[Generating synthetic EHR data using ABMs to evaluate ML robustness under MCIs]{Generating synthetic electronic health record data using agent-based models to evaluate machine learning robustness under mass casualty incidents}

\author{%
\Name{Roben {Delos Reyes}} \Email{rdelosreyes@student.unimelb.edu.au}\\
\addr School of Computing and Information Systems, The University of Melbourne, Parkville, Victoria, Australia
\AND
\Name{Daniel Capurro} \Email{dcapurro@unimelb.edu.au}\\
\addr School of Computing and Information Systems, The University of Melbourne, Parkville, Victoria, Australia\\
\addr Department of Medicine, The University of Melbourne, Parkville, Victoria, Australia
\AND
\Name{Nicholas Geard} \Email{ngeard@unimelb.edu.au}\\
\addr School of Computing and Information Systems, The University of Melbourne, Parkville, Victoria, Australia
}

\begin{document}

\maketitle

\begin{abstract}
Machine learning (ML) models in healthcare are typically evaluated using curated real-world electronic health record (EHR) data. A key limitation of such evaluations is that they may fail to assess the robustness of ML models to changes in the data at deployment, which is a common issue because EHR data used for ML model development cannot capture all such changes. Mass casualty incidents (MCIs) caused by disasters are critical instances where this will be an issue, as they induce rare, uncertain, and novel changes to routine system conditions. Because real-world EHR data from MCIs are often limited or unavailable, assessing ML robustness under such conditions before deployment remains challenging. Here, we propose an agent-based modelling approach for generating synthetic EHR data to evaluate the robustness of ML models under MCI scenarios. We use real-world EHR data to develop and calibrate an agent-based model (ABM) of an emergency department (ED) that explicitly models patient arrivals, resource capacity, and clinical workflow. By changing these system conditions to reflect plausible MCI scenarios, the ED model generates synthetic versions of the real-world EHR data that exhibit shifts in system behaviour. Using these synthetic data, we test ML models for predicting length of stay. We observed consistent declines in recall under MCI conditions relative to baseline system conditions, resulting in an increase in the number of patients with prolonged length of stay that were missed by the ML models. These results highlight the impact of changes in system conditions on patient outcomes, EHR data, and ML model performance. Our work establishes ABM-based synthetic EHR data generation as a proactive and systematic approach for evaluating the robustness of ML models under MCI or other system conditions not captured in real-world EHR data, supporting the safer and more effective deployment of ML models in healthcare systems. 

\end{abstract}

\paragraph*{Data and Code Availability}
This work uses the MIMIC-IV dataset, which is available on the PhysioNet repository \citep{Goldberger2000, Johnson2023b, Johnson2023c, johnsonMIMICIVFreelyAccessible2023}. The code used for this work is available at: \url{https://github.com/rddelosreyes/ed-abm-synthetic-ehr}.

\paragraph*{Institutional Review Board (IRB)}
This work does not require IRB approval. 

\section{Introduction}
\label{sec:intro}
Machine learning (ML) models are increasingly deployed in healthcare systems to support various aspects of clinical and operational decision making \citep{zhang2022shifting, Poon2025}. One important application of ML models is predicting patient outcomes, such as hospitalisation, inpatient mortality, readmission, and length of stay, as these predictions can inform decisions on patient care and resource management. Many ML models have been shown to make such predictions accurately when trained and tested on curated electronic health record (EHR) datasets \citep{Hilton2020, xieBenchmarkingEmergencyDepartment2022, Stone2022, Lee2024, farimani2024models}. Despite such promising results, a typical challenge with ML models is that they lack robustness to changes in the data caused by changes in the system in which they are used \citep{Zech2018, Nestor19, zhang2022shifting}. These changes in system conditions reflect real-world variations in patient populations, healthcare resources, and clinical practices present in the \textit{deployment} data that are not sufficiently captured in the \textit{development} data used to train and test the ML models \citep{finlaysonClinicianDatasetShift2021a}. Hence, for safety-critical applications such as healthcare, evaluating the robustness of ML models on data under various real-world system conditions during the ML model development phase is essential to ensure their safe and effective use in clinical practice \citep{Lekadir2025, balendran2025scoping}. 

Mass casualty incidents (MCIs) caused by disasters such as earthquakes and infectious disease outbreaks are critical instances in which the development data may not capture real-world situations, yet ML models must remain robust. When MCIs occur, healthcare systems experience sudden changes in demand and operating conditions \citep{carrington2021impact}. To evaluate the robustness of ML models under such conditions, EHR data reflecting system conditions and patient outcomes across plausible MCI scenarios are necessary. However, real-world EHR data needed for such robustness evaluation are often difficult to obtain for many reasons. For one, access to real-world EHR data is tightly regulated by national authorities and healthcare institutions due to the sensitive information they carry. Even when credentialed access is granted, components of the data are anonymised to comply with these regulations \citep{johnsonMIMICIVFreelyAccessible2023}. Furthermore, real-world EHR data capture only what was observed and recorded from reality, which may not include sufficient data for training and testing ML models under various MCI conditions. Thus, the standard process of evaluating ML model performance using a held-out test set from a given real-world EHR dataset is inherently limited in providing insights into the accuracy and robustness of ML models \citep{van2023can}. 

To address the limited availability of real-world data, synthetic EHR data have been used as complementary data sources \citep{gonzales2023synthetic}. Synthetic EHR data are artificially generated data based on real-world EHR data and medical knowledge using approaches like computational or generative ML models \citep{hernandez2022synthetic}. These synthetic EHR data generation approaches enable the training and testing of ML models for contexts where real-world EHR data are inadequate or unavailable. A key distinguishing feature of these approaches is that the generation process can be controlled to produce data with specific characteristics. Hence, plausible variations of real-world EHR data can be deliberately added in the synthetic EHR data \citep{van2023can}. The generated synthetic EHR data can then be utilised to test the robustness of ML models systematically against various robustness issues in healthcare \citep{finlaysonClinicianDatasetShift2021a, zhang2022shifting, balendran2025scoping}. 

However, many existing synthetic EHR data generation approaches lack the capability to explicitly model the system conditions of the healthcare system, where and when patients receive care. The current focus of existing approaches is to mimic the properties of real-world EHR data, such that the synthetic data represent patient characteristics and outcomes that are realistic but do not reveal real patient information \citep{yan2022multifaceted, budu2024evaluation}. In that process, they implicitly model the same system conditions in which the real-world EHR data were collected. Yet, such system conditions do not necessarily extend to MCI scenarios where demand and operating conditions change in uncertain ways \citep{carrington2021impact}. Moreover, patient outcomes are also affected by such changes in system conditions. For instance, crowding in the emergency department affects the quality of care, resulting in worse patient outcomes such as higher inpatient mortality, increased risk of readmission, and longer length of stay \citep{bernstein2009effect, morley2018emergency, bravata2021association, kadri2021association}. 

To be able to test how robust ML models are during MCIs, there remains a need to capture how system conditions and patient outcomes are reflected in real-world EHR data and control how they are integrated into their synthetic versions. Here, we investigate the use of agent-based models (ABMs) for generating synthetic EHR data to evaluate the robustness of ML models in predicting patient outcomes under MCI conditions. ABMs are mechanistic representations of real-world systems, where entities are modelled as autonomous agents that interact in a system environment according to defined behavioural rules \citep{bonabeau2002agent}.  In ABMs, system conditions of real-world systems can be explicitly modelled and simulated. Such an approach makes them particularly suited for investigating how interactions among agents and the environment affect emergent system properties, from understanding and mitigating disease outbreaks to analysing and optimising healthcare operations \citep{tracy2018agent, willem2017lessons, liu2017agent, delosreyes2026data}. ABMs have also been used to generate realistic synthetic populations for exploring how different policies and interventions affect demographic, social, and health patterns \citep{geard2013synthetic, predhumeau2023synthetic, von2025synthetic}. However, this capability of ABMs for synthetic data generation has not yet been fully explored and leveraged in existing literature on ML robustness.

\begin{figure*}[t!]
    \centering
    \includegraphics[width=\linewidth]{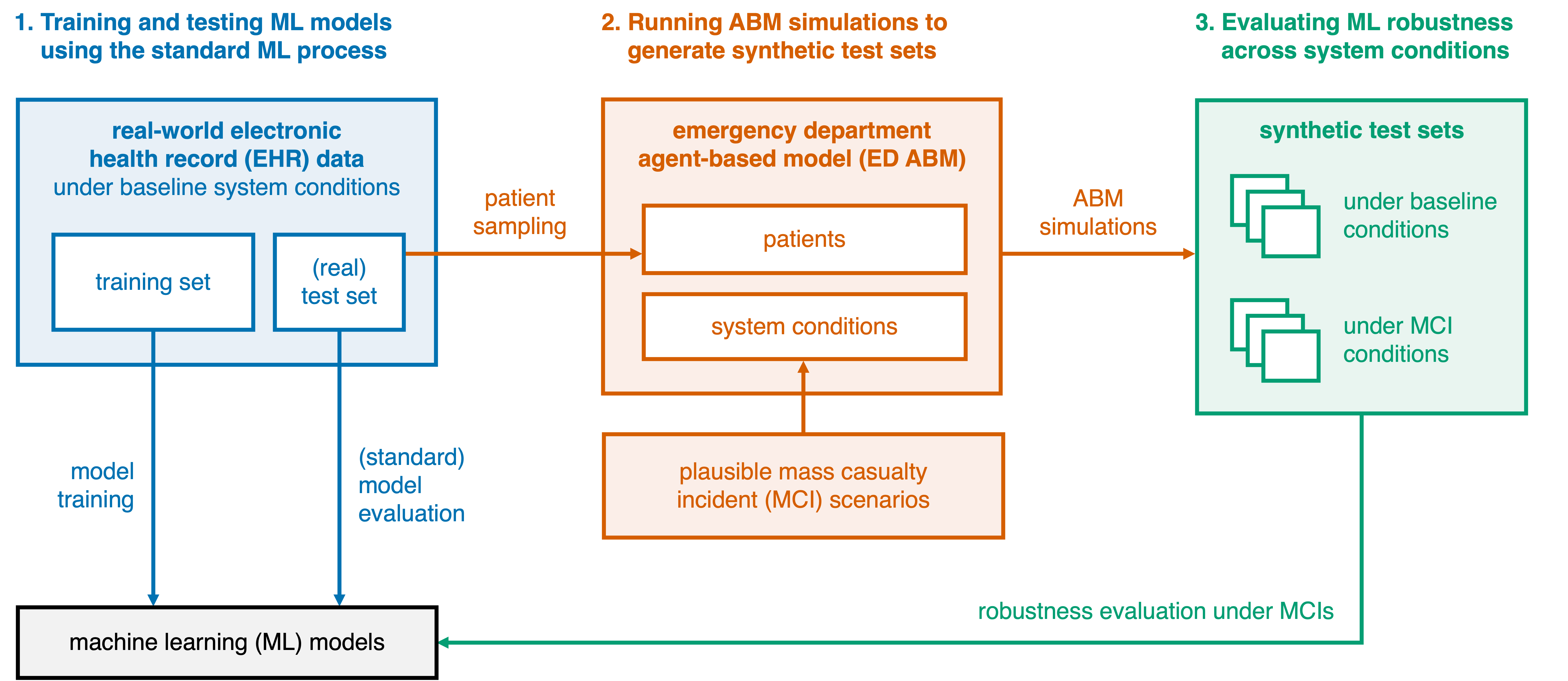}
    \caption{Schematic overview of our ABM-based approach for synthetic EHR generation and ML robustness evaluation under MCI scenarios.}
    \label{fig:fig1}
\end{figure*}

In this work, we use the emergency department (ED) as the operational setting. We first describe the standard ML process for training and testing an ML model for an ED task using real-world EHR data. We then present an ABM of an ED and demonstrate how patients captured in the real test set can be simulated in the ABM. Finally, we simulate these patients across various MCI scenarios and show how synthetic test sets generated from these ABM simulations can be used to evaluate the robustness of ML models under those conditions. We provide a schematic overview of our proposed approach in Figure \ref{fig:fig1}.

\section{Methods}
\label{sec:methods}
\subsection{Real-world EHR data}
We use real-world EHR data from the Medical Information Mart for Intensive Care (MIMIC) dataset to develop both the ML model (described in Section \ref{subsubsec:ml_model}) and the ABM of an ED (described in Section \ref{subsubsec:edabm}). MIMIC-IV, the latest version of MIMIC, provides credentialed access to de-identified data of patient admissions at the Beth Israel Deaconess Medical Center in Boston, Massachusetts \citep{Johnson2023c, johnsonMIMICIVFreelyAccessible2023}. We focus on ED admissions from the MIMIC-IV-ED module (version 2.2), which contains 425,087 records of ED stays from 2011 to 2019 \citep{Goldberger2000, Johnson2023b}. These records are associated with a unique stay identifier, which can be used to extract information on patients' demographics, conditions, and the various activities they undergo in the ED, including arrival, triage, vital sign checks, medicine dispensations and administrations, laboratory and imaging tests, and discharge. For simplicity, we consider each ED stay record to be associated with a unique patient $p$ in this work. 

\subsection{ML task and model} \label{subsubsec:ml_model}
We consider the ML task of predicting, at triage upon the patient's arrival in the ED, whether or not a patient will have a length of stay (LOS) in the ED beyond a prespecified duration \citep{farimani2024models}. LOS is a critical performance indicator in the ED that affects patient care and outcomes \citep{wiler2015emergency, vanbrabant2019simulation}. National health guidelines specify target LOS values for treating ED patients to ensure that they are receiving timely and adequate care (e.g., 4 hours in Australia and the UK \citep{vezyridis2014national, forero2019perceptions}). Accurate predictions of patients' LOS outcomes could thus support more effective resource utilisation and patient management.

To train and test an ML model for this task, we preprocessed the MIMIC-IV dataset following a benchmarking study of ML models for ED prediction tasks \citep{xieBenchmarkingEmergencyDepartment2022}. The preprocessed dataset made up the full dataset $\mathcal{D}$ used during the ML model development phase: 
\begin{equation}
    \mathcal{D}=\{(x_p,y_p)\}_{p=1}^{P},
\end{equation}
where $x_p$ is a vector of features characterising patient $p$, $y_p$ is an indicator of the outcome of patient $p$, and $P$ is the number of patients included in the full dataset $\mathcal{D}$. The feature vector $x_p$ encapsulates the patient-level features of patient $p$, which consists of age, vital signs, chief complaints, comorbidities, past admission counts, and acuity. The outcome label $y_p \in \{0,1\}$ indicates whether patient $p$ has an LOS greater than the LOS threshold $\ell$, where we set $\ell=4$. The full dataset $\mathcal{D}$ was then split into a training set $\mathcal{D}_{train}$ (80\%) and a test set $\mathcal{D}_{test}$ (20\%).

Given the feature vector $x_p$ of patient $p$, an ML model $f$ predicts the LOS outcome of patient $p$:
\begin{equation}
    \hat{y}_p = f(x_p; \theta_f),
\end{equation}
where $\hat{y}_p \in \{0,1\}$ is the predicted label of the outcome and $\theta_f$ denotes the parameters of the ML model $f$. The ML model's parameters $\theta_f$ are learned from the training set $\mathcal{D}_{train}$. The ML model's performance is measured based on the accuracy of the predicted outcome label $\hat{y}_p$ relative to the true outcome label $y_p$ for every patient $p$ in the test set $\mathcal{D}_{test}$. We trained and tested three common ML models: random forest, gradient boosting, and multilayer perceptron. These ML models are widely used for tabular data and demonstrate performance that is competitive with more recent architectures on such data \citep{Nestor19, Hilton2020, xieBenchmarkingEmergencyDepartment2022, van2023can, farimani2024models, Lee2024}. 

\subsection{ED ABM description} \label{subsubsec:edabm}
We adapt a previously published ABM of the ED at the Beth Israel Deaconess Medical Center, which was developed using the MIMIC-IV dataset \citep{Delosreyes2024}. For simplicity, we refer to it as ED ABM. Whereas the ML model described in Section \ref{subsubsec:ml_model} is designed to predict patients' LOS outcomes, this ED ABM is designed to generate synthetic EHR data for evaluating the robustness of the ML model in making those predictions under different system conditions of the ED. The ED ABM represents the ED environment and its resources, as well as the flow of patients through that environment. The state of the ED environment is defined by its \texttt{hourly\_arrival\_rate}, \texttt{bed\_capacity}, \texttt{clinician\_capacity}, \texttt{imaging\_capacity}, and \texttt{clinical\_workflow}. Every patient in this ED environment has a state variable for \texttt{age}, \texttt{vital\_signs}, \texttt{chief\_complaints}, \texttt{comorbidities}, \texttt{past\_admission\_counts}, \texttt{acuity}, \texttt{disposition}, \texttt{trajectory}, and \texttt{length\_of\_stay}.

The state of the ED ABM is updated at discrete time steps. At each time step, the following occurs. First, patient arrivals are generated stochastically following the prespecified \texttt{hourly\_arrival\_rate}. The state variables of each generated patient, except the \texttt{length\_of\_stay}, are specified according to the patient record sampled stochastically from the given dataset (explained in Section \ref{subsubsec:sampling}). Second, patients are assigned to beds based on their \texttt{acuity} and bed availability, with those of higher acuity given priority. Finally, patients in bed are treated. Once patients are discharged from the ED, their state variables are recorded and are then removed from the ED environment. 

Since resources are limited and shared among all patients in the ED environment, patients may have to wait for resources during their ED stay. This waiting time is not included in the patient's \texttt{trajectory}, which only encapsulates the execution time of each activity. Hence, the \texttt{length\_of\_stay}, which is the sum of all execution and waiting times, is a stochastic and emergent output of the ED ABM. 

\subsection{Sampling patients from the real test set} \label{subsubsec:sampling}
When patient arrivals are generated in the ED ABM, their state variables need to be initialised. Except for \texttt{length\_of\_stay} which is initialised to 0, we specify the values of these state variables based on patient records from the test set $\mathcal{D}_{test}$. This setup can be viewed as putting a patient from the test set $\mathcal{D}_{test}$ into the ED ABM. For every new patient $p$, we sample the patient's \texttt{acuity} and \texttt{disposition} based on observed frequencies from the test set $\mathcal{D}_{test}$. We then randomly sample with replacement a patient from the test set $\mathcal{D}_{test}$ with that acuity and disposition. The \texttt{age}, \texttt{vital\_signs}, \texttt{chief\_complaints}, \texttt{comorbidities}, and \texttt{past\_admission\_counts} of the new patient $p$ are then directly specified according to the sampled patient's ED stay record.

The \texttt{trajectory} of the new patient $p$ indicates the sequence and execution time of activities that the patient will undergo in the ED, starting with arrival in the ED and ending with discharge from the ED. Specifying the sequence and execution time of these activities requires further preprocessing of the MIMIC-IV dataset. To extract patient trajectories from the dataset, we obtained the event log of patients in the test set $\mathcal{D}_{test}$. An event log is a digital record of the activities that patients underwent in the system \citep{rojas2016process}. Each recorded event includes a patient identifier, the name of the activity, and the timestamp at which the activity was recorded into the EHR. By sorting each patient's recorded events chronologically, we can reconstruct the series of activities that patients underwent in the ED from arrival to discharge, and use that to specify the \texttt{trajectory} of the new patient $p$ in the ED ABM. Since the activity timestamps in the MIMIC-IV dataset only indicate the time at which an activity was recorded into the EHR, the recorded time duration between two activities may include both waiting and execution times for the latter activity. We removed the waiting times embedded in the recorded patient trajectories as detailed in Appendix \ref{apd:waiting_time}.

\setlength{\tabcolsep}{3pt}
\begin{table*}[t!]
\footnotesize
\centering
\begin{tabular}{@{}p{1.5cm}p{2cm}p{2.3cm}p{2.3cm}p{1.4cm}p{2.3cm}@{}}
\toprule
 & \multicolumn{3}{l}{\textbf{Population level}} & \multicolumn{2}{l}{\textbf{Patient level}}\\
 \cmidrule(lr){2-4} \cmidrule(lr){5-6}
 & \textbf{Real\newline median LOS} & \textbf{Simulated\newline median LOS} & \textbf{Wasserstein\newline distance} & \textbf{Coverage} & \textbf{Width}\\
\midrule
\multicolumn{6}{@{}l}{Acuity}\\
\hspace{5pt} 1 & 5.20 & 5.10 (4.69--5.50) & 0.86 (0.68--1.07) & 0.80 & 0.60 (0.40--0.89)\\
\hspace{5pt} 2 & 6.70 & 6.32 (6.12--6.56) & 0.86 (0.63--1.13) & 0.63 & 0.68 (0.45--0.98)\\
\hspace{5pt} 3 & 5.82 & 5.90 (5.60--6.28) & 0.65 (0.50--0.85) & 0.87 & 3.15 (2.14--4.26)\\
\hspace{5pt} 4 & 2.92 & 3.09 (2.73--3.48) & 0.73 (0.60--0.94) & 0.91 & 3.05 (2.04--4.12)\\
\hspace{5pt} 5 & 2.03 & 2.30 (1.43--3.71) & 1.42 (1.20--1.90) & 0.92 & 2.54 (1.45--3.75)\\
\midrule
\multicolumn{6}{@{}l}{Disposition}\\
\hspace{5pt} Home & 5.25 & 5.30 (5.05--5.60) & 0.77 (0.61--0.96) & 0.79 & 2.47 (0.95--3.83)\\
\hspace{5pt} Ward & 7.07 & 7.00 (6.76--7.28) & 0.55 (0.42--0.73) & 0.78 & 1.12 (0.59--2.91)\\
\hspace{5pt} ICU & 5.17 & 5.26 (4.92--5.60) & 0.71 (0.58--0.90) & 0.87 & 0.80 (0.50--1.32)\\
\midrule
Overall & 5.85 & 5.81 (5.60--6.07) & 0.59 (0.47--0.76) & 0.79 & 1.85 (0.72--3.50)\\
\bottomrule
\end{tabular}
\caption{Validation of the ED ABM under baseline system conditions. The ED ABM generated synthetic test sets that accurately reproduced the LOS distributions from the real test set across the overall patient population and different acuity and disposition groups. The simulated median LOS (in hours) closely matches the real median LOS, with small Wasserstein distances between distributions. At the patient level, simulated LOS intervals capture the real LOS value for 79\% of patients, with a median interval width of 1.85. Values in parentheses indicate the interquartile range across 1,000 simulation runs.}
\label{tab:tab2}
\end{table*}

\subsection{Generating synthetic test sets}
We used the ED ABM to generate synthetic EHR data under different system conditions. In formal terms, the ED ABM $g$ takes as input the \textit{real} test set $\mathcal{D}_{test}$ and outputs a \textit{synthetic} test set $\mathcal{D}_{syn}$:
\begin{equation}
    \mathcal{D}_{syn} = g(\mathcal{D}_{test}; \theta_g),
\end{equation}
where $\theta_g$ denotes the parameters of the ED environment that describe the system conditions in the ED. The ED environment's parameters $\theta_g$ include the hourly patient arrival rate, the resource capacity for beds, clinicians, and imaging equipment, and the ED workflow, which were all initially set based on previously calibrated values \citep{Delosreyes2024}. We note that these calibrated values represent the baseline system conditions of the ED, reflecting the system conditions from which data in the MIMIC-IV dataset were collected. By changing these parameter values, we can simulate the ED under different system conditions and generate a synthetic test set $\mathcal{D}_{syn}$ that reflects those conditions. 

Similar to the real test set $\mathcal{D}_{test}$, every patient $p$ included in the synthetic test set $\mathcal{D}_{syn}$ is also represented using the simulated patient-level feature vector $\bar{x}_p$ and the simulated outcome label $\bar{y}_p$:
\begin{equation}
    (\bar{x}_p, \bar{y}_p) \in \mathcal{D}_{syn}.
\end{equation}
As per the patient sampling described in Section \ref{subsubsec:sampling}, the simulated patient-level feature vector $\bar{x}_p \in \mathcal{D}_{syn}$ of patient $p$ is always similar to the real patient-level feature vector $x_p \in \mathcal{D}_{test}$. However, the simulated outcome label $\bar{y}_p \in \mathcal{D}_{syn}$ may differ from the true outcome label $y_p \in \mathcal{D}_{test}$ because patient $p$'s \texttt{length\_of\_stay} is also affected by the ED environment's parameters $\theta_g$. 

\subsection{Experimental setup}
We conducted two experiments to evaluate the utility of our proposed ABM-based approach for generating synthetic EHR data and evaluating ML robustness. Since the ED ABM is stochastic, we report results across 1,000 simulation runs, each run generating a synthetic test set. The first experiment evaluated how well the ED ABM could generate synthetic EHR data with LOS characteristics that match the MIMIC-IV dataset at the population and patient levels. For each simulation run $n$, we compared the simulated LOS values in the synthetic test set $\mathcal{D}_{syn}^n$ to the true LOS values in the real test set $\mathcal{D}_{test}$. At the population level, we examined the LOS distributions across the overall patient population and patient groups by acuity and disposition. At the patient level, we calculated coverage and width to examine whether each patient's LOS was reproduced correctly. Coverage measures the fraction of patients whose true LOS value falls within the interval of simulated LOS values generated across simulation runs, while width measures the average size of this interval across all patients \citep{van2023can}. In addition, we also compared the performance of the ML models trained on the real training set $\mathcal{D}_{train}$ when tested on the real test set $\mathcal{D}_{test}$ versus when tested on each synthetic test set $\mathcal{D}_{syn}^n$.

The second experiment assessed the robustness of ML models under MCI conditions using the synthetic test sets generated by the ED ABM. We simulated a 4-day MCI period, during which ED system conditions were systematically modified to reflect (1) an increase in patient arrivals, (2) a decrease in resource capacity, or (3) a delay in clinical workflow. Only patients who arrived in the ED during this time period were included in the analysis for both the baseline and MCI conditions. With 1,000 simulation runs under this setup, each patient was sampled an average of 9.4 times. We used the standard precision and recall metrics to measure ML model performance. To contextualise these ML model-based metrics in terms of patient and operational impact, we also calculated the number of patients with LOS $>$4 hours who were missed by the models per 100 ED stays. 

\setlength{\tabcolsep}{3pt}
\begin{table}[t!]
\footnotesize
\centering
\begin{tabular}{@{}p{3.3cm}p{2.2cm}p{2.1cm}@{}}
\toprule
& \textbf{Median LOS} & \textbf{LOS $>$4 hrs}\\
\midrule
\multicolumn{3}{@{}l}{Baseline conditions}\\
\hspace{5pt} Real test set & 5.85 & 0.74\\
\hspace{5pt} Synthetic test sets & 5.81 (5.60--6.07) & 0.75 (0.72--0.77)\\
\midrule
\multicolumn{3}{@{}l}{Synthetic MCI conditions: increase in arrivals}\\
\hspace{5pt} +5\% daily arrivals & 6.20 (5.88--6.56) & 0.78 (0.75--0.81)\\
\hspace{5pt} +10\% daily arrivals & 6.74 (6.28--7.29) & 0.81 (0.79--0.84)\\
\hspace{5pt} +15\% daily arrivals & 7.66 (6.97--8.60) & 0.85 (0.82--0.87)\\
\hspace{5pt} +20\% daily arrivals & 9.07 (7.97--10.49) & 0.87 (0.85--0.89)\\
\midrule
\multicolumn{3}{@{}l}{Synthetic MCI conditions: decrease in resources}\\
\hspace{5pt} -5\% clinicians & 6.02 (5.73--6.41) & 0.76 (0.73--0.79)\\
\hspace{5pt} -10\% clinicians & 6.30 (5.93--6.80) & 0.78 (0.75--0.82)\\
\hspace{5pt} -15\% clinicians & 7.39 (6.67--8.59) & 0.84 (0.81--0.88)\\
\hspace{5pt} -20\% clinicians & 8.72 (7.47--10.31) & 0.88 (0.84--0.91)\\
\midrule
\multicolumn{3}{@{}l}{Synthetic MCI conditions: delay in workflow}\\
\hspace{5pt} +5 mins for lab tests & 6.33 (6.04--6.65) & 0.79 (0.77-0.82)\\
\hspace{5pt} +10 mins for lab tests & 6.90 (6.58--7.29) & 0.83 (0.81--0.85)\\
\hspace{5pt} +15 mins for lab tests & 7.56 (7.17--8.08) & 0.86 (0.84--0.88)\\
\hspace{5pt} +20 mins for lab tests & 8.39 (7.83--9.17) & 0.89 (0.87--0.91)\\
\bottomrule
\end{tabular}
\caption{LOS characteristics of the real and synthetic test sets under baseline and MCI conditions. Median LOS (in hours) increases as system load increases (via an increase in arrivals, a decrease in resources, or a delay in workflow), as well as the fraction of patients with LOS $>$4 hours. Values are reported as median (interquartile range) across 1,000 simulation runs.} 
\label{tab:tab3}
\end{table}

\section{Results}
\label{sec:results}

\subsection{Baseline system conditions}
The ED ABM reproduced accurately the LOS values from the real test set under baseline conditions, as shown in Table \ref{tab:tab2}. For the overall patient population, the real median LOS is 5.85 hours, while the simulated median LOS is 5.81 hours. The median Wasserstein distance between the real and simulated LOS distributions is 0.59, indicating high similarity. Likewise, the true and simulated LOS distributions across different acuity and disposition groups have Wasserstein distances ranging from 0.55 to 1.42. The ED ABM also generated simulated LOS intervals that captured well the true LOS value of each patient. For the overall patient population, coverage is 79\% and width is 1.85. For the different acuity and patient groups, coverage ranges from 63\% to 92\%, while width ranges from 0.60 to 3.05. 

The three ML models also showed similar precision and recall on the real test set and synthetic test sets under baseline system conditions, as shown in Table \ref{tab:tab4}. On the real test set, model precision ranges from 0.78 to 0.79, while model recall ranges from 0.92 to 0.95. On the synthetic test sets, model precision ranges from 0.79 to 0.80, while model recall ranges from 0.91 to 0.95. Correspondingly, the number of missed patients with LOS $>$4 hours is 4--7 per 100 ED stays under baseline system conditions.

\subsection{Synthetic MCI conditions}
Under MCI conditions, LOS distributions in the synthetic test sets shifted relative to baseline system conditions, as shown in Table \ref{tab:tab3}. Increases in ED system load from MCIs, resulting from higher patient arrivals, reduced clinician capacity, and delays in laboratory workflow, led to longer LOS on average. Across the synthetic MCI scenarios considered, overall median LOS increased by 0.21--3.26 hours, while the fraction of patients with LOS $>$4 hours increased by 0.01--0.14. 

When evaluated on synthetic test sets generated under MCI conditions, all ML models exhibited a lack of robustness to shifts in LOS distributions arising from increases in system load, as shown in Table \ref{tab:tab4}. Although model precision increased by 0.01--0.12 relative to baseline, model recall consistently declined by 0.01--0.04. This reduction in recall translates into an additional 1--5 patients with LOS $>$4 hours missed per 100 ED stays, indicating a degradation in the ability of the ML models to predict patients at risk of staying longer in the ED during MCIs. 

\setlength{\tabcolsep}{2pt}
\begin{table*}[t!]
\footnotesize
\centering
\resizebox{\textwidth}{!}{
\begin{tabular}{@{}p{3.3cm}p{1.6cm}p{1.6cm}p{1.6cm}p{1.6cm}p{1.6cm}p{1.6cm}p{1.7cm}p{1.6cm}p{1.6cm}@{}}
\toprule
& \multicolumn{3}{l}{\textbf{Precision}} & \multicolumn{3}{l}{\textbf{Recall}} & \multicolumn{3}{p{5cm}}{\textbf{Missed patients with LOS $>$4 hrs\newline (per 100 ED stays)}}\\
\cmidrule(lr){2-4} \cmidrule(lr){5-7} \cmidrule(lr){8-10}
& \textbf{RF} & \textbf{GB} & \textbf{MLP} & \textbf{RF} & \textbf{GB} & \textbf{MLP} & \textbf{RF} & \textbf{GB} & \textbf{MLP}\\
\midrule
\multicolumn{10}{@{}l}{Baseline conditions}\\
\hspace{5pt} Real test set & 0.79 & 0.78 & 0.79 & 0.92 & 0.95 & 0.94 & 5.93 & 3.62 & 4.11\\
\hspace{5pt} Synthetic test sets & 0.80 $\pm$ 0.06 & 0.79 $\pm$ 0.06 & 0.79 $\pm$ 0.06 & 0.91 $\pm$ 0.03 & 0.95 $\pm$ 0.02 & 0.94 $\pm$ 0.03 & 6.64 $\pm$ 2.58 & 4.11 $\pm$ 2.06 & 4.70 $\pm$ 2.21\\
\midrule
\multicolumn{10}{@{}l}{Synthetic MCI conditions: increase in arrivals}\\
\hspace{5pt} +5\% daily arrivals & 0.82 $\pm$ 0.06 & 0.81 $\pm$ 0.06 & 0.81 $\pm$ 0.06 & 0.90 $\pm$ 0.03 & 0.94 $\pm$ 0.03 & 0.93 $\pm$ 0.03 & 7.77 $\pm$ 2.97 & 5.06 $\pm$ 2.46 & 5.71 $\pm$ 2.61\\
\hspace{5pt} +10\% daily arrivals & 0.84 $\pm$ 0.06 & 0.84 $\pm$ 0.06 & 0.84 $\pm$ 0.06 & 0.89 $\pm$ 0.03 & 0.92 $\pm$ 0.03 & 0.92 $\pm$ 0.03 & 9.15 $\pm$ 3.27 & 6.17 $\pm$ 2.68 & 6.93 $\pm$ 2.81\\
\hspace{5pt} +15\% daily arrivals & 0.87 $\pm$ 0.06 & 0.86 $\pm$ 0.06 & 0.86 $\pm$ 0.06 & 0.88 $\pm$ 0.03 & 0.91 $\pm$ 0.03 & 0.90 $\pm$ 0.03 & 10.40 $\pm$ 3.19 & 7.25 $\pm$ 2.62 & 8.06 $\pm$ 2.76\\
\hspace{5pt} +20\% daily arrivals & 0.88 $\pm$ 0.05 & 0.88 $\pm$ 0.05 & 0.88 $\pm$ 0.05 & 0.87 $\pm$ 0.03 & 0.91 $\pm$ 0.02 & 0.90 $\pm$ 0.03 & 11.24 $\pm$ 2.83 & 7.99 $\pm$ 2.34 & 8.85 $\pm$ 2.51\\
\midrule
\multicolumn{10}{@{}l}{Synthetic MCI conditions: decrease in resources}\\
\hspace{5pt} -5\% clinicians & 0.81 $\pm$ 0.07 & 0.80 $\pm$ 0.07 & 0.80 $\pm$ 0.07 & 0.91 $\pm$ 0.03 & 0.94 $\pm$ 0.03 & 0.93 $\pm$ 0.03 & 7.14 $\pm$ 2.98 & 4.55 $\pm$ 2.36 & 5.14 $\pm$ 2.55\\
\hspace{5pt} -10\% clinicians & 0.83 $\pm$ 0.07 & 0.82 $\pm$ 0.08 & 0.82 $\pm$ 0.08 & 0.90 $\pm$ 0.03 & 0.94 $\pm$ 0.03 & 0.93 $\pm$ 0.03 & 7.78 $\pm$ 3.35 & 5.03 $\pm$ 2.70 & 5.72 $\pm$ 2.88\\
\hspace{5pt} -15\% clinicians & 0.87 $\pm$ 0.08 & 0.87 $\pm$ 0.08 & 0.87 $\pm$ 0.08 & 0.89 $\pm$ 0.04 & 0.93 $\pm$ 0.03 & 0.92 $\pm$ 0.03 & 9.51 $\pm$ 3.98 & 6.37 $\pm$ 3.22 & 7.14 $\pm$ 3.48\\
\hspace{5pt} -20\% clinicians & 0.90 $\pm$ 0.08 & 0.89 $\pm$ 0.08 & 0.90 $\pm$ 0.08 & 0.88 $\pm$ 0.04 & 0.92 $\pm$ 0.03 & 0.91 $\pm$ 0.03 & 10.55 $\pm$ 3.86 & 7.20 $\pm$ 3.17 & 8.02 $\pm$ 3.36\\
\midrule
\multicolumn{10}{@{}l}{Synthetic MCI conditions: delay in workflow}\\
\hspace{5pt} +5 mins for lab tests & 0.84 $\pm$ 0.06 & 0.83 $\pm$ 0.06 & 0.83 $\pm$ 0.06 & 0.90 $\pm$ 0.03 & 0.94 $\pm$ 0.02 & 0.93 $\pm$ 0.03 & 7.62 $\pm$ 2.82 & 4.88 $\pm$ 2.21 & 5.51 $\pm$ 2.40\\
\hspace{5pt} +10 mins for lab tests & 0.87 $\pm$ 0.05 & 0.86 $\pm$ 0.05 & 0.86 $\pm$ 0.05 & 0.90 $\pm$ 0.03 & 0.93 $\pm$ 0.03 & 0.92 $\pm$ 0.03 & 8.60 $\pm$ 3.06 & 5.59 $\pm$ 2.49 & 6.32 $\pm$ 2.62\\
\hspace{5pt} +15 mins for lab tests & 0.89 $\pm$ 0.05 & 0.89 $\pm$ 0.05 & 0.89 $\pm$ 0.05 & 0.89 $\pm$ 0.03 & 0.93 $\pm$ 0.03 & 0.92 $\pm$ 0.03 & 9.64 $\pm$ 3.17 & 6.43 $\pm$ 2.61 & 7.22 $\pm$ 2.76\\
\hspace{5pt} +20 mins for lab tests & 0.92 $\pm$ 0.04 & 0.91 $\pm$ 0.04 & 0.91 $\pm$ 0.04 & 0.88 $\pm$ 0.03 & 0.92 $\pm$ 0.03 & 0.91 $\pm$ 0.03 & 10.68 $\pm$ 3.18 & 7.24 $\pm$ 2.61 & 8.10 $\pm$ 2.79\\
\bottomrule
\end{tabular}}
\caption{ML robustness evaluation results. As ED system load increases under MCI conditions, precision increases while recall decreases across all three ML models: random forest (RF), gradient boosting (GB), and multilayer perceptron (MLP). The number of missed patients with LOS $>$4 hours increases from 4--7 per 100 ED stays under baseline conditions to 5--11 per 100 ED stays under the synthetic MCI conditions. Results for each of the synthetic baseline and MCI conditions are shown as the mean $\pm$ 2 standard deviations across 1,000 test sets.}
\label{tab:tab4}
\end{table*}

\section{Discussion}
\label{sec:discussion}
Robustness evaluation should be a standard part of machine learning (ML) model evaluation before deployment of ML models in healthcare systems, determining when and how they can be safely and effectively used to support clinical and operational decisions \citep{finlaysonClinicianDatasetShift2021a, Lekadir2025, balendran2025scoping}. When evaluating ML models during the ML model development phase, the standard use of a held-out test set from a real-world electronic health record (EHR) dataset is inadequate for measuring the accuracy and robustness of these ML models in deployment because of its scarcity and potential data misalignment \citep{Zech2018, Nestor19, zhang2022shifting, van2023can}. Hence, relying solely on ML model performance based on evaluations on this held-out test set could lead to a retrospective identification of ML model vulnerabilities, posing risks to patients and healthcare providers.

Here, we focused on evaluating the robustness of ML models under mass casualty incidents (MCIs), wherein crowding resulting from changes in system conditions, such as increases in patient arrivals, decreases in resource capacity, and delays in clinical workflow, negatively affects patient outcomes, such as inpatient mortality and length of stay (LOS) \citep{bernstein2009effect, morley2018emergency, bravata2021association, kadri2021association}. Since MCIs are relatively rare, stochastic in nature, and potentially novel, real-world EHR data that adequately capture changes in system conditions and patient outcomes during MCIs are often limited or unavailable. This data limitation makes it difficult or impossible to evaluate ML robustness under MCI conditions. To enable this robustness evaluation, we presented an agent-based model (ABM) of an emergency department (ED) for generating synthetic EHR data during MCIs. 

By explicitly modelling system conditions, our ABM-based approach to generating synthetic EHR data can simulate realistic and interpretable changes to arrivals, resources, and workflow induced by MCIs. It accounts for how the impact of system conditions on patient outcomes is captured in real-world EHR data and enables control over how these system conditions and their impact are reflected in synthetic EHR data. Such capabilities are rarely considered or demonstrated in existing synthetic EHR data generation approaches \citep{hernandez2022synthetic, yan2022multifaceted, van2023can, budu2024evaluation}. Our work thus improves on existing approaches to synthetic EHR data generation and ML model evaluation by enabling the stress testing of ML models under MCI and other system conditions expected in deployment but not sufficiently captured in the development data. The system conditions in the ABM can be readily adapted to consider other scenarios. For instance, simultaneous changes in arrivals, resources, and workflow can also be explored to simulate more complex MCI scenarios. Since increases in system load due to each of these changes led to longer LOS, as shown in Table \ref{tab:tab3}, such simultaneous changes would be expected to further increase system load and result in similar findings. 

In practice, our approach could be applied to EHR data from other healthcare systems to generate synthetic data relevant to the specific real-world scenarios they confront. These synthetic data would support proactive and systematic evaluation of the robustness of ML models under various real-world system conditions, informing ML model deployment and indicating when further improvement is required. While there will always be uncertainty about how the synthetic EHR data compare to real-world EHR data during MCIs, they can provide insights into the robustness of ML models in healthcare that are not obtainable with real-world EHR data alone nor with synthetic EHR data generated using existing approaches.

Many ML models designed to predict patient outcomes often rely solely on patient-level features, such as patients' demographics, past medical encounters, and current health conditions \citep{Nestor19, xieBenchmarkingEmergencyDepartment2022, Stone2022, van2023can, Lee2024, farimani2024models}. In part, this oversight is due to the abstraction of system conditions from EHR datasets, such as publicly available benchmarks like MIMIC, to protect the privacy of patients and healthcare providers \citep{johnsonMIMICIVFreelyAccessible2023}. However, in the simulation experiments, many patients had longer LOS under MCI conditions with higher system load than under baseline system conditions, regardless of their patient-level features, as exhibited in Table \ref{tab:tab3}. Consequently, ML models trained solely on patient-level features, following common practice, failed to predict LOS under those MCI conditions as accurately as under baseline system conditions, as demonstrated in Table \ref{tab:tab4}. Hence, when using real-world EHR data or generating synthetic EHR data for training and testing ML models during the ML model development phase, how system conditions and their impact are encapsulated in the data should also be explicitly considered (e.g., by incorporating system-level features) to assess whether ML models are robust under expected real-world system conditions, such as in situations like MCIs when they are needed more. 

We introduced ABMs as an approach for explicitly modelling and simulating these system conditions and their impact on patient outcomes. The use of ABMs to support the training and testing of ML models is underexplored, despite them being widely used to inform decision making across healthcare systems \citep{tracy2018agent}. Here, we showed that an ABM can reproduce and extend real-world EHR data, as shown in Tables \ref{tab:tab2} and \ref{tab:tab3}, respectively. Further, we demonstrated that the generated synthetic EHR data can be used to evaluate the robustness of ML models under MCI conditions, as shown in Table \ref{tab:tab4}. Our work can also be used as a reference for many other potential applications of ABMs for ML research and education. First, ABMs could also be used to generate synthetic data for evaluating other ML robustness issues in healthcare \citep{finlaysonClinicianDatasetShift2021a, zhang2022shifting, balendran2025scoping}. Second, they could complement existing synthetic EHR data generation approaches designed for generating completely synthetic patients \citep{yan2022multifaceted, van2023can, budu2024evaluation}. Third, they could augment system conditions in public EHR benchmarks like MIMIC, which are often anonymised for privacy reasons, to enable further studies on the effect of system-level features on ML model performance \citep{johnsonMIMICIVFreelyAccessible2023}. Fourth, they could also be used to generate synthetic EHR data for training ML models to enhance their robustness. Finally, ABMs could also serve as clinical environment simulators, in which ML models are embedded within ABMs to enable the dynamic evaluation of their impact on system behaviour over time \citep{luo2026clinical}. 

Our work has several limitations. First, we assumed that recorded time durations between ED activities that deviate from an average value include waiting times, which may not necessarily be the case. Despite this, we validated that this choice generated synthetic EHR data with similar LOS characteristics to the real-world EHR data. Further, this removal process of waiting times could be useful in cases when waiting times are not explicitly captured in the EHR data. Second, ML model evaluation under MCI conditions is based only on synthetic scenarios and would thus require EHR data during real-world MCIs for further validation. However, the emergent data shifts observed under these MCI conditions (i.e., increasing LOS with increasing system load) align with expectations and are supported by extensive literature on ED crowding \citep{bernstein2009effect, morley2018emergency}. Finally, we focused on modelling and predicting LOS only. Future work could extend our approach to evaluate the robustness of ML models for predicting other patient outcomes such as hospitalisation, inpatient mortality, and readmission.

\section{Conclusion}
\label{sec:conclusion}
In summary, we developed an ED ABM that can generate synthetic EHR data during MCIs. We then demonstrated how the synthetic EHR data can be used to evaluate the robustness of ML models under various MCI conditions. Our results showed that ML models for predicting LOS outcomes lack robustness to changes in system conditions induced by MCIs. Our work extends existing synthetic EHR data generation approaches by enabling the explicit modelling of system conditions. It also extends the standard ML model evaluation process by offering a proactive and systematic approach to evaluating the robustness of ML models under MCI conditions before deployment. Our work thus helps ensure that ML models can be safely and effectively used in clinical practice. Furthermore, it provides a basis for novel and relevant applications of ABMs in ML for health.

\section*{Author contributions: CRediT}
\textbf{Roben Delos Reyes}: Conceptualization, Methodology, Software, Validation, Formal Analysis, Investigation, Data Curation, Writing - Original Draft, Writing - Review \& Editing, Visualization. \textbf{Daniel Capurro}: Conceptualization, Methodology, Writing - Review \& Editing, Supervision. \textbf{Nicholas Geard}: Conceptualization, Methodology, Writing - Review \& Editing, Supervision.

\section*{Funding sources}
This work was supported by the Melbourne Research Scholarship provided by the University of Melbourne.

\acks{
The experiments conducted in this work were supported by the resources provided by the University of Melbourne’s Research Computing Services and the Petascale Campus Initiative.}

\bibliography{chil-sample}

\newpage
\appendix
\section{Removing waiting times from recorded patient trajectories}\label{apd:waiting_time}

We assumed that the time interval between the timestamps of two consecutive activities, say, an activity $A$ followed by an activity $B$, includes not only the execution time of activity $B$ but also potentially a waiting time for activity $B$. Formally, we express the recorded time duration $\tau_{AB}$ to include both the waiting time $w_{AB}$ for and execution time $e_{AB}$ of an activity $B$, from an activity $A$:
\begin{equation}
    \tau_{AB} = w_{AB} + e_{AB}.
\end{equation}

To remove the waiting times, we performed temporal conformance checking wherein we checked whether the recorded time duration $\tau_{AB}$ deviated from the average time duration between activities $A$ and $B$ \citep{stertz2020temporal}. First, we identified the average time duration between every activity $A$ and $B$ by calculating its median $Mdn_{AB}$ and mean absolute deviation $MAD_{AB}$ across all patient trajectories in the test set $\mathcal{D}_{test}$. Second, for every activity $A$ to $B$ in a patient's recorded trajectory, we calculated the modified z-score of the recorded time duration $\tau_{AB}$:
\begin{equation}
    z = \frac{0.6745(\tau_{AB}-Mdn_{AB})}{MAD_{AB}}.
\end{equation}
Every recorded time duration $\tau_{AB}$ that has a modified z-score greater than a prespecified z-score threshold $k$ is considered to be a \textit{temporal deviation} and thus includes a waiting time. On the contrary, if the modified z-score is less than or equal to the z-score threshold $k$, then the recorded time duration $\tau_{AB}$ is considered to be in conformance to the expected duration and thus has no waiting time ($\tau_{AB} = e_{AB}$).

We then removed the waiting times from the recorded time durations recognised as temporal deviations. We generated the updated time duration $\tau'_{AB}$ of every activity $A$ to $B$ in the recorded trajectories as follows:
\begin{equation}
    \tau'_{AB} = 
    \begin{cases}
        \tau_{AB} & \text{if } z \leq k\\
        Mdn_{AB} + k(1.4826*MAD_{AB}) & \text{if } z > k.
    \end{cases}
\end{equation}
We used these updated time durations to specify the execution time of activities in the \texttt{trajectory} of the new patient $p$ in the ED ABM. In the results shown in Section \ref{sec:results}, the z-score threshold $k=3$. This removal of waiting times from recorded patient trajectories was determined through calibration experiments.

\end{document}